\documentclass[runningheads]{llncs}

\usepackage{eccv}

\definecolor{myred1}{HTML}{FF6464} 
\definecolor{myred2}{HTML}{FFB4B4} 
\definecolor{myred3}{HTML}{FFE6E6} 

\usepackage{eccvabbrv}

\usepackage{graphicx}
\usepackage{booktabs}
\usepackage{multirow} 
\usepackage{bytefield}
\usepackage{amsmath}
\usepackage[table]{xcolor}
\usepackage{booktabs}
\usepackage{siunitx} 

\usepackage{bbm}

\usepackage[accsupp]{axessibility}  

\usepackage{hyperref}

\usepackage{orcidlink}

\begin{document}

\title{CGS-SLAM: Collaborative Gaussian Splatting based SLAM for Multi-Agent Reconstruction} 

\titlerunning{CGS-SLAM: Collaborative Gaussian Splatting based SLAM}

\author{Jean-Daniel de Ambrogi\inst{1,2}\orcidlink{0009-0003-9368-1545} \and
Aladine Chetouani\inst{1}\orcidlink{0000-0002-2066-4707} \and
Vincent Nguyen\inst{3}\orcidlink{0000-0003-2271-6918} \and
Aurélien Chateigner \inst{2}\orcidlink{0000-0003-4738-2096}}

\authorrunning{J.D.~de Ambrogi et al.}

\institute{ Université Sorbonne Paris Nord, L2TI, UR 3043, F-93430, Villetaneuse, France \and
SAS IMPACT, Orléans, France\and
Université d’Orléans, INSA CVL, LIFO, UR 4022, Orléans, France\\
}

\maketitle

\begin{abstract}
Recent advances in SLAM have leveraged 3DGS for photorealistic reconstruction and novel view synthesis. However, most methods rely on RGB-D input, which is unavailable on consumer-grade smartphones, and few integrate 3DGS within a collaborative framework. Therefore, we present CGS-SLAM, a hybrid decentralized/centralized system enabling multi-agent 3DGS SLAM using only RGB and inertial data. Each agent performs local tracking with inertial data as a motion prior and reconstructs a scaled map using a metric monocular depth estimator (Depth Pro). Keyframe encodings are shared among agents, enabling dynamic keyframing in regions of spatial overlaps with other agents, enhancing submap alignment. Afterwards, a central server aligns submaps using VGGT as a view alignment model. This bidirectional communication keeps communication cost low during mapping and global reconstruction in difficult GNSS-denied environments. Experiments on multiple datasets demonstrate competitive tracking performance, improved rendering quality over state-of-the-art methods, and accurate submap alignment.
\end{abstract}

\section{Introduction}
In high-stakes operational scenarios such as hostage rescue, disaster response, or reconnaissance missions, law enforcement and military units are frequently required to navigate unfamiliar, GNSS\footnote{Global Navigation Satellite System}-denied environments under time-critical constraints. These operations often involve rapid deployment across confined or structurally complex spaces, with minimal prior knowledge of the terrain and coordination among multiple, concurrently operating teams.  
To enable effective decision-making, there is a need for a rapid mapping system able to fuse the local contributions of each team into a globally consistent, shared representation.

This problem falls within the domain of Simultaneous Localization and Mapping (SLAM), which enables reconstruction of spatial environments while concurrently localizing the exploratory agent within them. However, most existing SLAM systems rely on explicit depth priors\cite{mappingLandscapeSLAM}, either from dedicated depth sensors or via dense structure-from-motion (SfM) pipelines\cite{sfm,sfm_review}, to compensate for the inherent ambiguity of monocular vision. Although effective, these approaches often incur significant computational overhead, rendering them impractical for deployment on resource-constrained platforms, limiting live operation, since they require both time and parallax-inducing motion at initialization.

In practice, the current electronic equipment available to tactical units is ruggedized consumer-grade smartphones, limited to RGB cameras and inertial measurement units (IMUs). They typically do not provide any depth measure through LiDAR or Time-of-Flight (ToF) cameras. This sensor constraint narrows the design space for viable SLAM solutions, as it precludes the use of active depth sensors.

Furthermore, many existing systems output point clouds or mesh-based reconstructions, both of which are sensitive to trajectory coverage. In scenarios where exploration is sparse, these representations often exhibit significant topological incompleteness and are hard to interpret. Recent advances in 3D Gaussian Splatting (3DGS)\cite{kerbl3Dgaussians} offer an alternative. While it is not immune to sparse coverage either, its continuous, opacity-weighted formulation does not require a topological decision in unobserved regions, unlike an explicit surface, and with its photorealistic representations and fast rendering it allows easier interpretation.

We present CGS-SLAM, a new Visual-Inertial SLAM (VI-SLAM) approach targeting this specific sensor-constrained regime. Our method reconstructs dense, continuous local maps on each agent, which are then aligned on a server using a sparse reconstruction model to produce a globally consistent shared map enabling real-time situational awareness for task force teams, even in GNSS-denied, resource-constrained environments.

Our main contributions, forming a global SLAM solution, are: 
\begin{itemize}
    \item A monocular inertial local SLAM system, leveraging a metric monocular depth estimation model, producing metrically scaled 3DGS maps from the very first frame. 
    \item A light-weight collaborative framework for constrained networks, offering dynamic keyframing capabilities for enhanced sub-map alignment.
    \item A rapid submap alignment process, leveraging VGGT to produce an immediate pose estimation of local submaps. 
\end{itemize}

\section{Related Work}
\label{sec:related_work}
\subsection{Gaussian Splatting}
Accurate, robust, and semantically meaningful environmental representation lies at the core of SLAM systems. Researchers thus increasingly turned to neural representations as the foundational primitive for scene reconstruction. 3D Gaussian Splatting (3DGS), introduced by Kerbl \etal\cite{kerbl3Dgaussians}, has rapidly transformed the field of 3D scene representation. It has found applications in mainstream industries such as film making~\cite{dune}~or medical education \cite{anatomy} due to its ability to provide photorealistic rendering at high frame rates. Several recent works \cite{CG-SLAM,CoSurfGS,gaussiansplattingSLAM,GS-SLAM,MM3DGS,MAGIC-SLAM,macego3d} have shown the feasibility of using 3DGS as a primary mapping representation in SLAM systems. However, most of these methods are heavily dependent on depth priors. While Sun \etal\cite{MM3DGS} proposed an RGB-only SLAM configuration, their method relies on relative monocular depth estimation, which provides inconsistent estimations during exploration and thus can lead to catastrophic map collapse. In our work, motivated by the need for a real-time rendering method, we adopt 3DGS as our core 3D representation. Further, we replace depth sensors with a recent state-of-the-art metric monocular depth estimation model (MMDE), Depth Pro \cite{mldepthpro}, that provides consistent at scale depth predictions, enabling accurate 3DGS reconstruction without LiDAR, ToF or stereo cameras.
\subsection{Monocular SLAM}
Given that most consumer devices lack the capability to capture explicit depth priors, SLAM research has increasingly focused on monocular-only solutions \cite{AbaspurKazerouni2022, mappingLandscapeSLAM, Li2022}. Early approaches relied on feature-based visual methods, extracting and matching hand-crafted image features for pose estimation and mapping \cite{ORBSLAM3_TRO, ptam}. However, these methods often exhibited limited robustness in environments with sparse texture, varying illumination, or rapid camera motion, which adversely affected depth estimation accuracy under such conditions. 
Subsequent works\cite{lift_slam,dxslam} have replaced traditional feature extractors with deep learning-based alternatives. Some have bypassed explicit feature matching altogether, regressing camera poses directly using end-to-end neural networks. More recently, advances in monocular depth estimation (MDE) have been integrated into SLAM pipelines\cite{MM3DGS,motslam,10356045}, enabling the replacement of external depth priors with depth maps predicted by specialized deep models. While these approaches have demonstrated success in producing geometrically coherent reconstructions, they typically suffer from a lack of global scale consistency, as MDE models inherently provide only relative depth with respect to the input image. Furthermore, dynamic scenes, illumination changes, or motion-induced artifacts can cause temporal inconsistencies in predicted depth maps across frames, leading to erroneous inputs for the mapping module and compromising overall trajectory and structure estimation. To mitigate these issues, we rely on the metric scale of our MMDE, thus improving the temporal consistency of depth prediction during exploration. 
\subsection{Collaborative SLAM}
Collaborative or multi-agent SLAM, which enables multiple agents to simultaneously localize and map an environment through information exchange, has attracted increasing interest but remains a challenging and relatively underdeveloped field compared to single-agent SLAM, due to technical complexities such as inter-agent data association, communication constraints, and map merging\cite{collaborativeSLAMreview}. While collaboration has been investigated using traditional map representations such as point clouds or meshes, few approaches utilize neural representations. Hu \etal\cite{cp-slam} enabled collaborative mapping through a neural point-based representation anchored to camera keyframes, fused across agents by a distributed-to-centralized learning scheme. Yugay \etal\cite{MAGIC-SLAM} proposed a multi-agent SLAM system based on Gaussian Splatting, achieving high-fidelity results with multiple agents; however it relies on RGB-D inputs requiring LiDAR sensors. Gao \etal\cite{CoSurfGS} introduced a collaboration strategy for wide-area scene capture using Unmanned Aerial Vehicles (UAV), focusing on stitching together local maps from each UAV. Nonetheless, this method assumes the availability of GNSS data and does not address the case where it is absent, thus leaving the SLAM tracking problem unaddressed. Finally, Xu \etal\cite{macego3d} proposed an RGB-D-based method leveraging Multi-Agent Gaussian Consensus achieving global consistency. However, it relies on intensive communication between agents, aiming for global consensus on overlapping areas and continuously refining Gaussians. Moreover it aligns submaps through Mahalanobis distance on point clouds, thus relying heavily on accurate mapping produced from RGB-D input.
\section{Method}
\begin{figure}[h]
    \centering
    \includegraphics[width=1\linewidth]{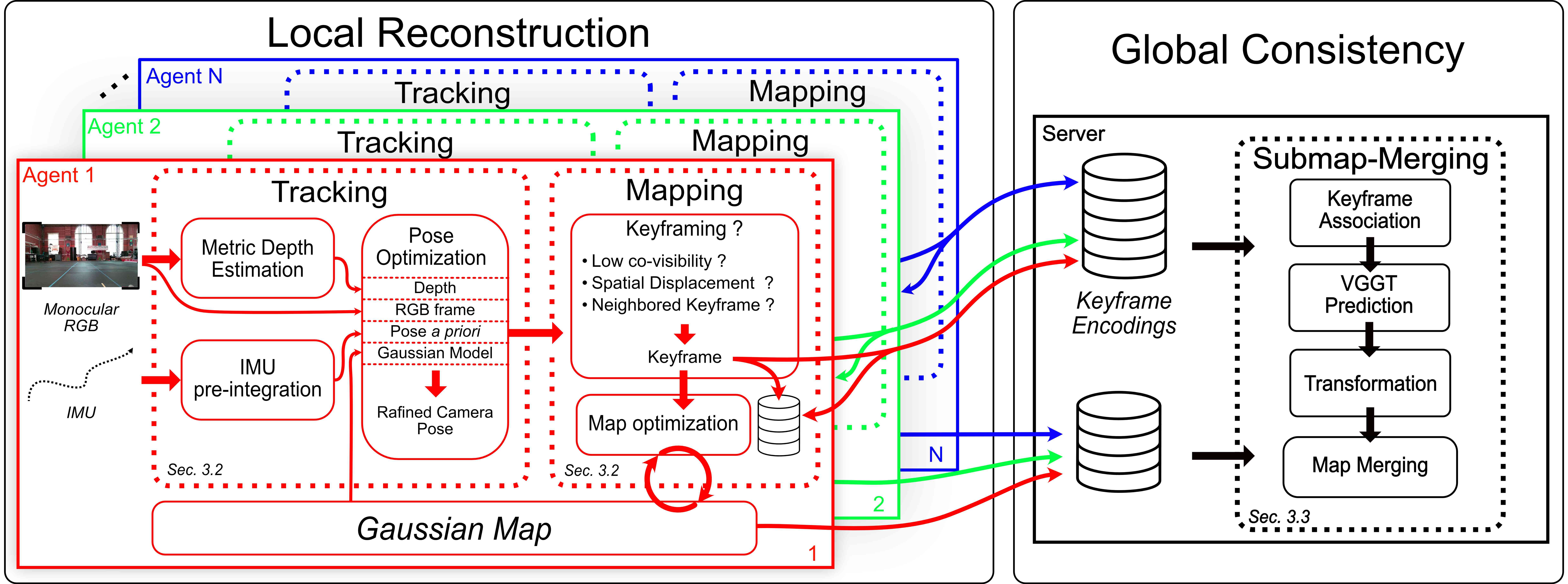}
    \caption{\textbf{The global architecture of our framework.} Our method allows multiple agents to evolve into an unknown environment and locally reconstruct it. The keyframes encodings are sent to the server, which broadcast them to all agents. Afterwards all agents share their local maps to the server which merge them into a single global map.}
    \label{fig:architecture}
\end{figure}
We propose a hybrid decentralized–centralized architecture for light-weight collaborative Gaussian‑Splatting SLAM. An overview of our framework is illustrated by \cref{fig:architecture}. Each agent independently reconstructs a local map from its own limited sensor suite, employing a decentralized approach while periodically exchanging keyframe encodings through a central server, allowing a dynamic keyframing process. After all agents finish their exploration, the centralized server fuses the individual submaps into a globally consistent map. 

In the sequel, we first describe the environment perception method, through IMU and depth estimation. Then we explore our framework for the local reconstruction method. Finally we detail the global consistency approach through our alignment pipeline.

\subsection{Environment Perception}

\subsubsection{Sensor Fusion}\label{sensorfusion}

In a monocular setup, where no reliable depth information is available, a basic approach  is to assume constant velocity between consecutive frames\cite{Tourani2022,Gu2023,ekfslam,vislampriors,visual_priors} as an initial motion estimate. Methods then rely on visual clues to optimize the pose of the camera through dense structure from motion\cite{ORBSLAM3_TRO,monocular_slam_review}. However, this assumption is not robust and can introduce significant errors in the predicted pose, especially in the presence of rapid camera movements or when the frame rate is low. Visual Inertial SLAM approaches \cite{OKVIS2, ORBSLAM3_TRO, rovio, VIGS-SLAM} have been proposed to tackle these limitations by leveraging IMU data as a more accurate \textit{a priori}. We adopt this approach with the aim of using all sensors that can help our method to have a better perception of the scene, and then estimate an accurate prior.

Thus, before optimizing the camera pose for each new frame, we compute an initial pose estimate from IMU preintegration.
The method to integrate and fuse these sensor signals is detailed in the supplementary material.
Considering that the IMU sensors have a significantly higher sensing frequency than the RGB camera, we chain the relative transformation throughout all IMU measurement up to the timestamps of the new tracked RGB frame.
Finally the camera pose $\mathbf{P}_j$ at the timestamp of the new RGB frame is  estimated from the previous camera pose $\mathbf{P}_i$ through:
\begin{equation}\label{imu_integration}
\mathbf{P}_j =\Bigg(\displaystyle\prod_{t=i}^{j} {}^{t-1}_{t} \mathbf{T}_C \Bigg)\mathbf{P}_{i}
\end{equation}
Where $i$ denotes the timestamp of the last tracked camera frame, \(j\) represents the timestamps of the new tracked frame and ${}^{t-1}_{t} \mathbf{T}_C$ is the translation matrix between the two consecutive timestamps $t$ and $t-1$.

\subsubsection{Monocular Depth Estimation}\label{depthestimation}

In order to circumvent the need for scale recovery through post-hoc alignment or global optimization, we employ in our experiments a metric-MDE (MMDE). The metrical dimension of the model also ensures temporal consistency across frames where relative depth estimates may drift due to exposure variations or motion blur. At each keyframe (selection detailed in \cref{densification}), depth is inferred using this model and subsequently integrated into the SLAM pipeline as a proxy for true metric depth. This approach eliminates the reliance on batch-based methods that require accumulating multiple keyframes to perform deep bundle adjustment, thereby enabling truly online operation from the very first frame while maintaining geometric consistency.

\subsection{Local Reconstruction}

Each agent constructs its local 3D map using synchronized RGB video and IMU data. As described in \cref{sensorfusion}, we fuse these complementary signals from the IMU to initialize the camera pose $P_i$, which serves as a coarse pose prior during the tracking phase, described in the next subsection. During this phase, we align in a frame-to-model manner the tracked pose to the Gaussian map $G$. Afterwards a frame is promoted to keyframe status if it fulfills certain conditions detailed below and then contributes to the map $G$ through projections of pixels using an MMDE described in \cref{depthestimation}. We then optimize the map $G$ through \cref{loss_mapping}. Recent keyframes are stored in a sliding window for local bundle adjustment, and jointly optimized to preserve geometric consistency over time. 

Upon completion of local exploration, the agent transmits its local map, including keyframes, Gaussian parameters, and camera intrinsics, to the server for global reconstruction described in \cref{align}
\subsubsection{Tracking}\label{tracking}

Since tracking quality depends on the geometric accuracy of the map, we complement the photometric term with a depth term. The metric model bounds the variation of the predicted depth range across consecutive estimates, but does not remove it entirely; we therefore compute the depth loss using the Pearson correlation coefficient (\cref{loss_depth}) between the estimated depth map \( D_e \) and the rendered depth map \( D_r \). Sun \etal \cite{MM3DGS} reported the benefit of this approach over an L1 depth term; we adopt their approach.
\begin{equation}\label{loss_depth}
    \mathcal{L}_{\text{depth}} = 1 - \rho(D_e, D_r) = 1 - \frac{\text{Cov}(D_e, D_r)}{\sqrt{\text{Var}(D_e) \text{Var}(D_r)}}.
\end{equation}
During Tracking, the Gaussian map is held constant, and the camera pose is optimized using the following composite loss function:
\begin{equation}\label{loss_tracking}
    \mathcal{L}_{\text{track}} = \operatorname{Mask}_{O(G, T_c)} \left( \mathcal{L}_{\text{photo}} + \lambda_D \mathcal{L}_{\text{depth}} \right)\text{, } \text{ and } 0 \leq \lambda_D \leq 1
\end{equation}
where \(\mathcal{L}_{\text{photo}}\) is the L1 loss between the ground truth and the rendered image and \( \operatorname{Mask}_{O(G, T_c)} \) is a masking function defined as:
\begin{equation}
    \operatorname{Mask}_{O(G, T_c)} =
    \begin{cases}
        1, & \text{if } O(G, T_c) > 0.99, \\
        0, & \text{otherwise}.
    \end{cases}
\end{equation}
This function keeps pixels whose computed opacity $O(G, T_c)$ is greater than 0.99, a threshold we found effective to retain only high opacity regions during the optimization. 

Finally, tracking is not applied to the first frame, as no Gaussian map has yet been constructed. Instead, the identity matrix is used as the initial camera transformation. The pose tracking is thus relative to the position of the first frame.

\subsubsection{Mapping}
\paragraph{Initialization}

Upon the first frame, pixels are back-projected into 3D space using the initial depth estimate provided by the metric depth model. This yields a scale-consistent initialization of the scene geometry, anchoring the mapping process to a metric reference frame. The 3DGS technique used in our method is detailed in the supplementary material.
\paragraph{Keyframe Selection}\label{densification}

The Gaussian representation is incrementally densified as new keyframes are selected based on a multi-criterion information gain metric. Specifically, it is designed to balance geometric coverage, temporal consistency, and computational efficiency. A new keyframe is instantiated according to the following conditions, checked in the following order: 

\begin{itemize}
\item \textbf{Spatial displacement}: To enforce temporal sparsity and avoid excessive keyframe insertion during slow or static motion, we enforce a minimum frame interval between consecutive keyframes. Thus, if the frame index is different from the last keyframe index by less than $s$ frames, the frame is discarded. Otherwise, a new keyframe is triggered if the current frame index exceeds the index of the last keyframe by at least $k$ frames. This ensures that keyframes are spaced sufficiently apart in time, reducing computational overhead while maintaining sufficient temporal resolution for robust tracking and mapping. 
\item \textbf{Sufficient information gain}: To ensure that each new keyframe contributes meaningful reconstruction information, we evaluate the spatial overlap between the current view and the most recent keyframe. Specifically, we render the depth map from the latest keyframe’s pose using the current Gaussian representation and identify pixels with valid depth and high silhouette confidence (i.e., presence mask > 0.99). These pixels are back-projected into 3D space to form a point cloud, which is then tested for co-visibility under the current camera pose. If the fraction of co-visible Gaussians does not exceed a predefined threshold (here 90\%), the frame is promoted to keyframe status. 
\item \textbf{Dynamic keyframing}: Throughout exploration, each agent continuously receives, through the server, the keyframe encodings computed by the other agents (\cref{global}). When an agent traverses an area previously mapped by another one, the encoding of the incoming frame exhibits high
similarity to one of these received encodings. We leverage this similarity as an indicator of spatial overlap between local reconstructions, and dynamically increase the keyframe sampling density within the overlapping region to enhance global consistency during subsequent map fusion.
\end{itemize} 

\paragraph{Map Optimization}
The previously selected keyframes are retained within a sliding window of fixed extent, over which periodic map optimization is performed to mitigate catastrophic forgetting and preserve geometric consistency throughout the reconstruction.

With \(\mathcal{L}_{\text{photo}}\) the same loss as in \cref{loss_tracking} and \(\mathcal{L}_{\text{D-SSIM}} = \frac{1 - \text{SSIM}}{2}\), the optimization of the Gaussian representation is governed by the following composite loss function: 
\begin{multline}\label{loss_mapping}
    \mathcal{L}_{\text{mapping}} = \lambda_{C}\mathcal{L}_{\text{photo}} 
    + \lambda_{S}\mathcal{L}_{\text{D-SSIM}} 
    + \lambda_D \mathcal{L}_{\text{depth}}, \\
    \quad \text{with } \lambda_{C}, \lambda_{S}, \lambda_D \in (0;1)
    \text{ and } \lambda_{C} + \lambda_{S} + \lambda_D = 1
\end{multline}
Following \cite{kerbl3Dgaussians,MM3DGS}, we employ the structural dissimilarity index (D-SSIM) \cite{SSIM} as a perceptual regularization term to enhance rendering fidelity.

Finally, a local bundle adjustment is performed over the keyframes retained in the sliding window. This step jointly refines camera poses and Gaussian parameters to ensure geometric coherence, effectively adapting the trajectory and map structure to the evolving scene representation.

\subsection{Global Consistency}\label{global}
\subsubsection{Dynamic Keyframing.} 

Submap alignment relies on the existence of keyframe pairs observing a common region. Our agents actively bias their keyframe distribution towards such regions: each agent transmits the encoding
of every new keyframe to the server, which caches them and broadcasts them to all other connected agents. Each agent matches incoming encodings against its own local keyframe database, and a match triggers the densification policy of \cref{densification}. Only fixed-size descriptors are exchanged; the mechanism therefore does not require a shared coordinate frame. The effect on alignment is twofold. Overlapping regions yield more candidate pairs for the keyframe association step described below, and each candidate pair comes with a denser set of temporal neighbours, which better conditions the subsequent pose refinement. The message protocol and the payload format are detailed in the supplementary material.

\subsubsection{Submap Alignment} \label{align}

Upon receiving the local map of each client (including RGB keyframes), the server initiates global map reconstruction by identifying pairs of co-visible keyframes across distinct local maps. These corresponding keyframe pairs are established using NetVLAD\cite{netvlad} visual descriptors. Keyframes with similar NetVLAD embeddings are considered candidates for spatial correspondence, under the assumption that visually similar frames are likely to capture overlapping regions of the environment. If no similar frames are found between two submaps, these submaps are not merged.

For each candidate pair $(\text{client}_1, \text{client}_2)$, we extract the keyframe images pairs, as well as several neighboring keyframes (from our experiments we choose 6 per submap, 3 before and after the keyframe). We feed them into VGGT\cite{wang2025vggt} to obtain camera extrinsic matrices $\mathbf{E}_i \in \mathbb{R}^{4\times4}$, which map camera coordinates to a specific and temporary VGGT world frame. These extrinsics are used with the locally tracked poses $\mathbf{T}^{S_i}_{C_i}$ (camera $\to$ submap) to compute an estimation of an initial rigid transformation $\mathbf{T}^{S_1}_{S_2}$ that aligns the second submap into the first’s coordinate frame. This transformation estimation is detailed in the supplementary material.

Since sub-maps from our front-end are metric-scale while VGGT operates up to relative scale, $\mathbf{T}^{S_1}_{S_2}$ contains the relative scale difference. We decompose $\mathbf{T}^{S_1}_{S_2}$ into rotation $\mathbf{R}$, translation $\mathbf{t}$ and scale $\mathbf{s}$, which is discarded, retaining only the rigid transformation $(\mathbf{R}, \mathbf{t}) \in SE(3)$.

This initial transformation is then refined via gradient-based optimization over the full set of neighboring keyframes. The optimization minimizes the geometric error between corresponding camera poses under the current transformation:

\begin{equation}
\mathcal{L}(\mathbf{T}) = \sum_{j} \left\| \mathbf{T} \cdot \mathbf{P}_j - \mathbf{Q}_j \right\|^2,
\end{equation}

where $\mathbf{P}_j$ are the source poses (transformed VGGT poses) and $\mathbf{Q}_j$ are the target poses (SLAM poses). The optimization is performed over the 6-DoF parameters of $\mathbf{T}$. Finally, this pose-based estimate is refined by a residual rigid correction, initialized at identity and optimized with Adam through the differentiable rasterizer, which minimizes the depth discrepancy between the two sub-maps rendered from a common keyframe pose.

\paragraph{Merging Submaps}
Once the optimal transformation $\mathbf{T}^{S_1}_{S_2}$ is computed, the second client’s Gaussian model is transformed into the first’s coordinate frame via:

\begin{equation}
\mathbf{G}_2' = \mathbf{T}^{S_1}_{S_2} \cdot \mathbf{G}_2,
\end{equation}

Applying \(\mathbf{T}^{S_1}_{S_2}\) to every Gaussian of client 2 brings its whole sub-map into the reference frame of client 1.

Once aligned, each local Gaussian map is incrementally merged into the global map using the computed transformation without further refinement to keep the process fast, at the cost of duplicated Gaussians on overlapping regions. All submaps are merged into the reference frame of the first completed and received local map, and the process is repeated as new local maps are received.

\newcommand{\best}[1]{\cellcolor{myred1}{#1}}
\newcommand{\secondbest}[1]{\cellcolor{myred2}{#1}}
\newcommand{\third}[1]{\cellcolor{myred3}{#1}}

\section{Experiments}
\subsection{Datasets}
A wide range of datasets have been proposed to benchmark SLAM\cite{6907054,tumvi,tumvie,ARKitscenes,arkittrack,lamar,indoormcd,isprs,Kaveti2023,MM3DGS,sturm12iros}, each tailored to specific sensor configurations, motion profiles, and environmental conditions. However, only a limited subset of these datasets provide synchronized monocular video sequences paired with IMU measurements under handheld operation in indoor environments. Multi-agent datasets, such as the ones derived from  \cite{replica19arxiv,shotton2013scene}, do not provide synced IMU data. Testing our pipeline on these datasets with constant velocity assumption would contaminate the evaluation, and using depth data would invalidate our lightweight-sensing promise. To evaluate our method, we select two datasets providing IMU, \textbf{TUM RGB-D Dataset} \cite{sturm12iros} and \textbf{UT-MM Dataset} \cite{MM3DGS}. While \textbf{TUM RGB-D Dataset} offers RGB-D sequences with ground-truth trajectories and accelerometer data, \textbf{UT-MM Dataset} consists of multiple RGB-D sequences with synchronized IMU sensors. Although it does not provide explicit multi-agent trajectories, all sequences were recorded within the same physical environment, enabling synthetic multi-agent evaluation.

\subsection{Experimental Set-up}
All experiments were conducted on a consumer-grade laptop, a Lenovo ThinkPad T15g Gen 1 equipped with an NVIDIA GeForce RTX 2080 Max-Q GPU and an Intel® Core™ i7-10750H CPU. This hardware configuration achieves a sustained frame rate of 10 FPS during tracking and mapping tasks. The client-side tracking and mapping module operates within approximately 1 GB of VRAM, making it suitable for deployment on portable workstations.

We use Depth Pro \cite{mldepthpro} as our MMDE for its depth quality. But this choice was at the cost of offloading it to the centralized server as it is too heavy for our laptop. 
Our pipeline is nonetheless MMDE-agnostic: we additionally evaluated Depth Anything v3, which replaces both the MMDE and the view alignment model for the submap alignment (VGGT). But it did not match the combination of Depth Pro and VGGT, results are available in the supplementary material.

The centralized server was equipped with an NVIDIA RTX 4090 GPU, producing depth estimation with Depth Pro at $\sim$169 ms per keyframe. The keyframe-encoding payload (described in the supplementary material) was 65 KB at 1--2 keyframes/s on UT-MM, yielding $<$130 KB/s uplink per agent, making it suitable for exchanges under degraded network conditions. The final maps were between 25 and 75 MB. 
\subsection{Performance of Monocular Single Agent Exploration}
\begin{figure}
  \includegraphics[width=\linewidth]{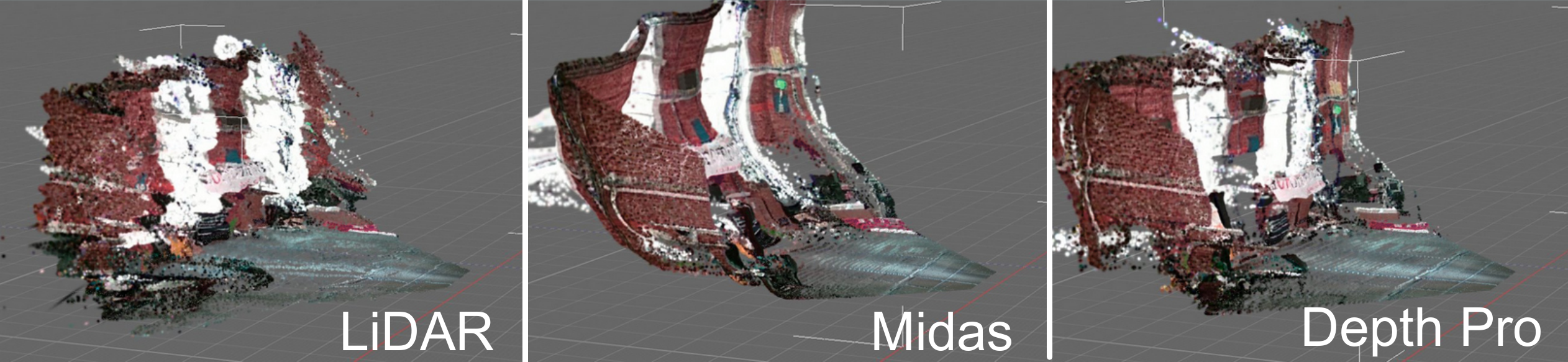}
  \caption{Comparison of the rendering between LiDAR (left), MiDaS\cite{midas} depth estimator (middle) and Depth-Pro (right) based Gaussian maps on scene \textit{fast-straight} from UT-MM. Our method exhibits reduced linear distortion along the horizontal axis. Even when compared against LiDAR measurements, our method demonstrates superior consistency against sensor artifacts found in the left panel.}
  \label{fig:fast-straight-three-comparaison}
\end{figure}

\begin{figure}
    \centering
    \includegraphics[width=\linewidth]{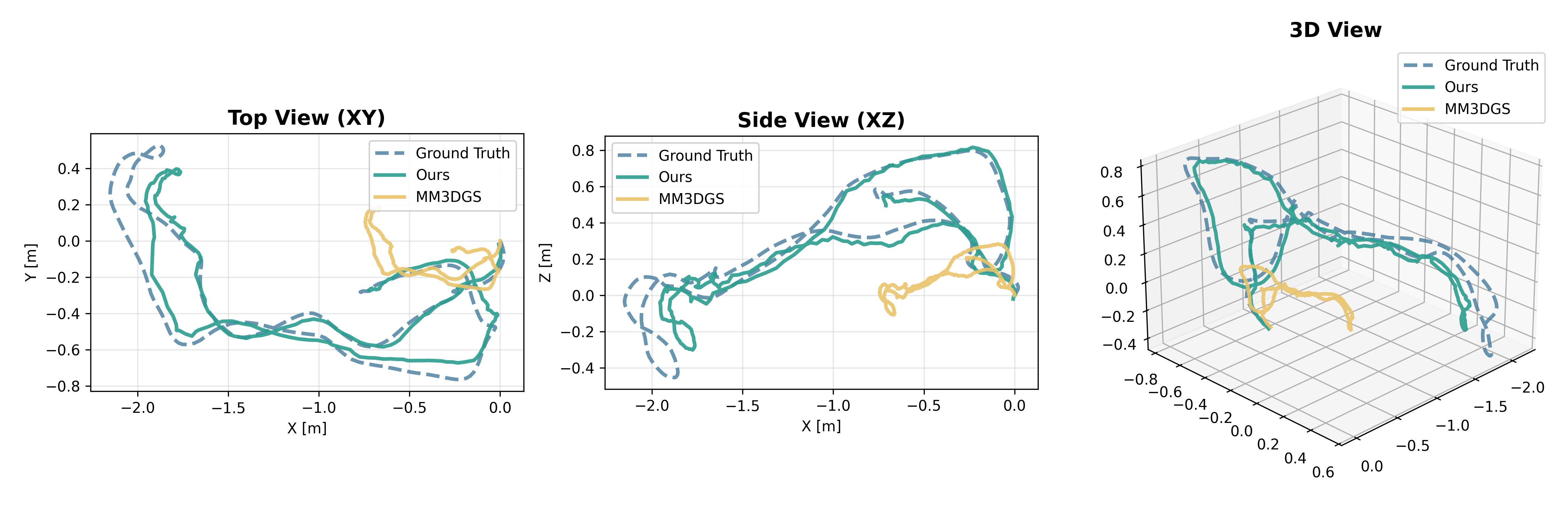}
    \caption{Raw trajectory output (TUM, fr1/desk), \ie when results are not rotated, scaled nor translated to fit at best the ground truth, as usually done to evaluate ATE-RMSE. We only align the origin. Our method offers a trajectory almost at scale with the ground truth, while MM3DGS is way off with an overall size of a third of the GT.}
    \label{fig:rawtrajectory}
\end{figure}

To evaluate the impact of integrating a metric depth model as a substitute for direct depth sensing, we conduct several evaluation comparisons. First, as illustrated in \cref{fig:fast-straight-three-comparaison}, the adoption of a modern metric depth estimation model yields  qualitative improvements over data from consumer grade LiDAR which can be noisy.
Furthermore, the metric depth model enables an accurate scaling of the reconstructed scene. In \cref{fig:rawtrajectory} we compare the raw trajectory estimation between our method and MM3DGS, we can see that our method accurately predicts the scale of the trajectory, and is close to the GT.
We also analyze the trajectory estimation performance of our framework under monocular conditions, specifically when no initial depth measurement is available to calibrate the scale of the first depth estimate. In such scenarios, the system must infer scale purely from visual cues and the predicted depth prior. 

\begin{figure}
    \centering
    \includegraphics[width=1\linewidth]{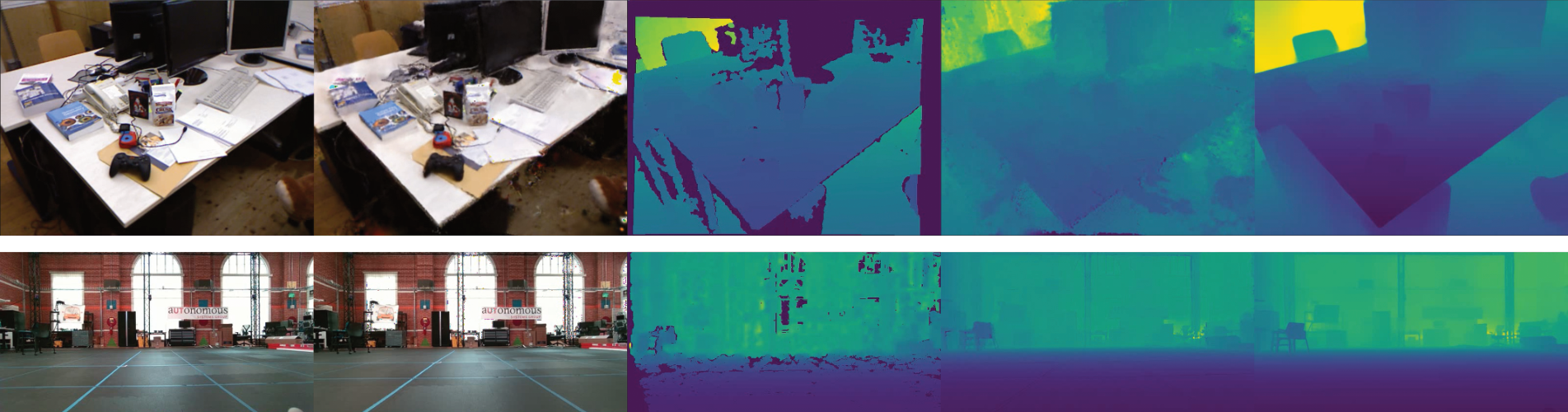}
    \caption{From left to right: RGB GT, RGB rendering, Depth GT, Depth rendering and Depth predicted from fr1/desk (TUM) and Fast-Straight (UT-MM) scenes with our method on monocular settings. We achieve clear rendering fidelity and spatial consistency with the GT.}
    \label{fig:renderer}
\end{figure}

 \Cref{tab:monoculatATERMSE} presents a comparative evaluation of ATE RMSE across multiple sequences under both RGBD, as reference, and monocular configurations. The results indicate that our approach achieves competitive tracking accuracy compared to methods that utilize LiDAR or RGBD inputs. In particular, we show competitive results with Magic-SLAM, the SOTA RGBD-based 3DGS SLAM method. This shows that a metric MDE can serve as a viable and effective replacement for direct depth sensing. 
\begin{table}
  \centering
    \resizebox{\columnwidth}{!}{%

  \setlength{\tabcolsep}{6pt}
  \begin{tabular}{llcccccccccc}
    \toprule
    Method & Map & Setup & fr1/desk & fr1/desk2  & fr1/room & fr2/xyz & fr3/office & Avg\\
        \midrule
    iMAP\cite{imap} & Neural &RGBD & 4.9* & -- & -- &  2.0*  & 58* & --\\
    NiceSLAM\cite{nice-slam} & Neural &RGBD & 4.26* & 4.99* & 34.49* &  31.73* & 30* & 21.09\\
    CP-SLAM\cite{cp-slam} & Neural &RGBD & 7.84* & -- & -- & 3.93* & -- & --\\
    Go-SLAM\cite{go-slam} & Neural &RGB & 1.6* & 2.8*& 5.2*  & 1.0* & 2.23 & 2.56\\
    \midrule
    GS-SLAM\cite{GS-SLAM} & 3DGS &RGBD & \best{3.3*} & -- & -- &  1.3* & -- & -- \\ 
    MAGIC-SLAM\cite{MAGIC-SLAM} & 3DGS &RGBD & 4.71 & \best{3.91} & \third{14.45} &  \secondbest{1.15} & \secondbest{15.64} & \secondbest{7.97}\\
    Mac-Ego3D\cite{macego3d} & 3DGS &RGBD & 25.53 & 30.81 & 22.50 & 1.39 & \best{11.62} & \third{18.37} \\
    MM3DGS\cite{MM3DGS} & 3DGS &RGB & \secondbest{3.35} & \third{6.54} & \secondbest{5.8}  & \third{1.24 } & 231.42 & {49.67}\\
    Ours & 3DGS &RGB + IMU & \third{4.2} & \secondbest{}{5.12} & \best{5.3} & \best{1.14} & \third{20.42} & \best{7.24}\\

    \bottomrule
  \end{tabular}
  }
  \caption{Tracking results against SOTA neural methods (top) and 3DGS based methods (bottom) on TUM Dataset (ATE-RMSE~$\downarrow$cm). Results marked with * are provided by the original paper. Our method does not use depth measurements and still performs as well as the RGBD-based 3DGS SLAM methods. The \colorbox{myred1}{first}, \colorbox{myred2}{second}, and \colorbox{myred3}{third} ranks are highlighted accordingly on 3DGS based SLAM; Go-SLAM provides the best results on all scenes. }
  \label{tab:monoculatATERMSE}
\end{table}
To assess the visual fidelity of the reconstructed scenes, we compute the PSNR, SSIM\cite{SSIM} and LPIPS\cite{LPIPS} between the original keyframes and their corresponding renderings generated by the final Gaussian map, evaluated at the estimated camera poses. As summarized in \cref{tab:psnr}, our approach, which employs a metric MDE, consistently outperforms the monocular Gaussian splatting based SLAM system MM3DGS on all evaluated sequences and matches the RGB-D baselines on most, Magic-SLAM remaining ahead on fr1/room. \cref{fig:renderer} illustrates this rendering quality through visual examples on two scenes.
\begin{table}[h]
  \centering
  
  \resizebox{\columnwidth}{!}{%
  \setlength{\tabcolsep}{1.5pt}
  \begin{tabular}{lcccccc|cccccc}
  \toprule
 & \multicolumn{3}{c}{Magic-SLAM} & \multicolumn{3}{c}{MAC-Ego3D} & \multicolumn{3}{c}{MM3DGS} & \multicolumn{3}{c}{Ours} \\
  \cmidrule(lr){2-4} \cmidrule(lr){5-7} \cmidrule(lr){8-10} \cmidrule(lr){11-13}
 Scene&PSNR&SSIM&LPIPS&PSNR&SSIM&LPIPS&PSNR&SSIM&LPIPS&PSNR&SSIM&LPIPS\\
    \midrule
    Square-1 & \secondbest{18.46} & \best{0.77} & \best{0.37} & \best{20.60} & 0.18 & 0.76 & 16.06 & \third{0.57} & \third{0.44} & \third{18.13} & \secondbest{0.64} & \secondbest{0.43}\\
    Square-2 & -- & -- & -- & \best{20.84} & \third{0.17} & \third{0.78} & \third{17.50} & \secondbest{0.60} & \secondbest{0.42} & \secondbest{18.08} & \best{0.62} & \best{0.39}\\
    Ego-centric & \third{19.10} & \third{0.79} & \third{0.30} & 14.38 & 0.39 & 0.61 & \secondbest{22.81} & \secondbest{0.82} & \secondbest{0.22} & \best{23.34} & \best{0.86} & \best{0.20}\\
    Fast-Straight & \secondbest{22.45} & \best{0.86} & \secondbest{0.28} & 21.04 & 0.20 & 0.77 & \third{21.58} & \third{0.78} & \third{0.29} & \best{24.51} & \secondbest{0.82} & \best{0.26}\\
    \midrule
    fr1/desk & \secondbest{16.57} & \secondbest{0.68} & \third{0.47} & \third{15.66} & \third{0.67} & \best{0.38} & 14.73 & 0.56 & 0.60 & \best{19.44} & \best{0.70} & \best{0.38}\\
    fr1/desk2 & \secondbest{17.04} & \best{0.70} & \third{0.46} & 14.66 & \secondbest{0.66} & \best{0.38} & \third{14.88} & 0.57 & 0.51 & \best{18.53} & \third{0.62} & \secondbest{0.43}\\
    fr1/room & \secondbest{15.56} & \secondbest{0.62} & \third{0.52} & \best{16.67} & \best{0.68} & \best{0.36} & 13.80 & 0.48 & \third{0.52} & \third{15.41} & \third{0.52} & \secondbest{0.48}\\
    fr2/xyz & \third{19.56} & \secondbest{0.79} & 0.33 & 19.46 & 0.74 & \third{0.26} & \secondbest{23.09} & \third{0.78} & \secondbest{0.24} & \best{23.24} & \best{0.84} & \best{0.20}\\
    fr3/office & \third{16.94} & \secondbest{0.71} & \third{0.44} & \best{19.23} & \best{0.74} & \best{0.28} & 10.71 & 0.37 & 0.60 & \secondbest{18.39} & \third{0.67} & \secondbest{0.42}\\
    \bottomrule
    \end{tabular}
    }
   \caption{Rendering comparison on TUM and UT-MM datasets. PSNR\(\uparrow\), SSIM\(\uparrow\) and LPIPS\(\downarrow\). Magic-SLAM and MAC-Ego3D are RGBD, while MM3DGS and ours are RGB-only. We perform better than MM3DGS on all scenes, and achieve similar quality with the RGBD methods. Magic-SLAM failed on Square-2 scene.}
   \label{tab:psnr}
\end{table}
Importantly, this improvement in rendering fidelity is achieved without compromising tracking accuracy. As previously established, our method maintains competitive trajectory estimation performance relative to other neural depth based approaches. Thus, the integration of a metric depth model delivers a dual benefit: it preserves the robustness of monocular SLAM while elevating the quality of the final reconstruction to a level that rivals that of sensor aided systems. 

We analyze the impact of IMU pre-integration by comparing against a constant velocity baseline, where angular velocity and linear acceleration are kept constant from the previous tracking step. \Cref{tab:imu} reports both the Absolute Trajectory Error (ATE) and PSNR metrics.

\begin{table}[h]
  \centering
 \resizebox{\columnwidth}{!}{
  \begin{tabular}{lcccccccccc}
    \toprule
    \multicolumn{1}{c}{Method} & \multicolumn{2}{c}{Square1} & \multicolumn{2}{c}{Ego-centric} & \multicolumn{2}{c}{Square2} & \multicolumn{2}{c}{Fast-straight} & \multicolumn{2}{c}{Avg} \\
    \cmidrule(lr){2-3} \cmidrule(lr){4-5} \cmidrule(lr){6-7} \cmidrule(lr){8-9} \cmidrule(lr){10-11}
      & \multicolumn{1}{c}{ATE} & \multicolumn{1}{c}{PSNR} & \multicolumn{1}{c}{ATE} & \multicolumn{1}{c}{PSNR} & \multicolumn{1}{c}{ATE} & \multicolumn{1}{c}{PSNR} & \multicolumn{1}{c}{ATE} & \multicolumn{1}{c}{PSNR} & \multicolumn{1}{c}{ATE} & \multicolumn{1}{c}{PSNR}\\
    \midrule
    MM3DGS (CV)  & 59.48 & 16.0 & \secondbest{4.09} & \third{23.15} & \third{67.20} & \third{17.51} & 25.78 & \third{21.71} & 39.14 & \third{19.59}\\
    Ours (CV)  & \third{55.43} & \secondbest{17.0} & 5.02 & \best{23.52} & \secondbest{62.84} & \secondbest{17.96} & \secondbest{12.85} & \secondbest{21.78} & \third{34.03} & \secondbest{20.06}\\
        \midrule
    MM3DGS (IMU) & \secondbest{47.05} & \third{16.06} & \best{3.52} & 22.81 & 68.50 & 17.50 & \third{16.78} & 21.58 & \secondbest{33.96} & 19.49\\
    Ours (IMU) & \best{42.33} & \best{18.13} & \third{4.26} & \secondbest{23.34} & \best{55.24} & \best{18.08} & \best{6.07} & \best{24.51} & \best{26.98} & \best{21.02}\\
    \bottomrule
  \end{tabular}
  }
  \caption{Contribution of the IMU pre-integration compared to the Constant Velocity (CV) Assumption during the tracking phase. All methods are in monocular settings, our method outperforms MM3DGS in most cases. ATE RMSE is in cm and PSNR is in dB. The \colorbox{myred1}{first}, \colorbox{myred2}{second}, and \colorbox{myred3}{third} ranks are highlighted accordingly.}
  \label{tab:imu}
\end{table}

IMU integration provides substantial tracking improvements on challenging sequences with rapid motions. On Fast-straight, our IMU-enabled method reduces ATE from 12.85 cm to 6.07 cm, a 53\% reduction while simultaneously improving PSNR by 2.73 dB. Similarly, on Square-1, ATE decreases from 55.43 cm to 42.33 cm.

Notably, our method with constant velocity (ATE: 12.85 cm) outperforms MM3DGS with IMU integration (ATE: 16.78 cm) on Fast-straight, suggesting that the proposed spatial alignment and map initialization contribute independently to tracking accuracy. The combination of IMU pre-integration with our initialization achieves the best overall performance, reducing ATE relative to the constant velocity baseline while establishing good rendering quality across all sequences.

\subsection{Performance on Global Reconstruction}

To validate our collaborative back-end, we evaluate the accuracy of submap alignment using the UT-MM dataset\cite{MM3DGS} as a multi-agent benchmark. All sequences in the dataset were captured within the same physical environment, the Anna Hiss Gymnasium at the University of Texas. This spatial consistency allows us to treat the dataset as a multi agent benchmark. We can evaluate trajectory alignment across independently captured runs within a shared coordinate frame. Our results indicate that trajectory alignment succeeds in the majority of cases, validating the feasibility of multi agent reconstruction under these constraints.
\begin{figure}[h]
    \centering
    \includegraphics[width=1\linewidth]{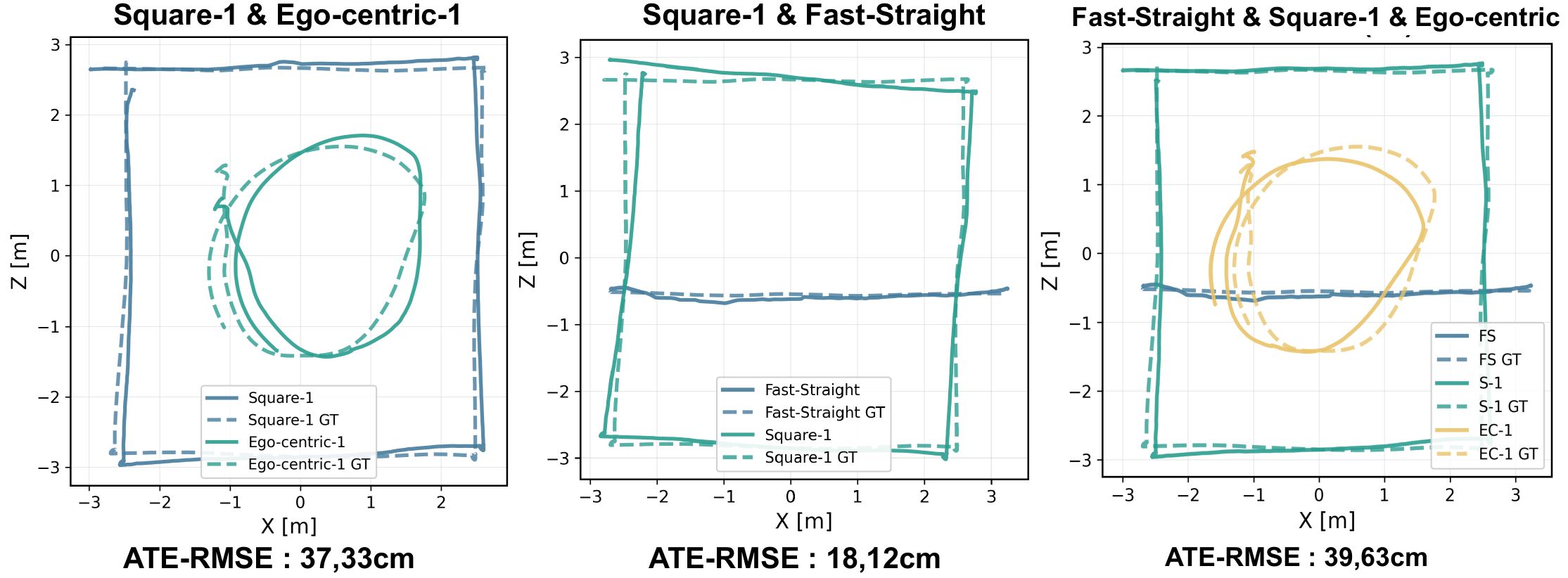}
    \caption{Submap alignment comparison on different scenes association from the UT-MM dataset. Each trajectory is plotted after transformed by the server sub-map alignment process. Our method achieves sub-meter spatial alignment with the ground truth (translation error < 0.40 m), and can work with multiple ( > 2) agents exploring the same scene.}
    \label{fig:global_alignement}
\end{figure}
As illustrated in \cref{fig:global_alignement}, our framework successfully aligns trajectories even with significant heading differences. It is worth noting that our method completes the alignment process in less than 10 seconds, allowing to obtain a global map almost directly upon receiving final submap. For example with Square-1 \& Fast-straight: $<$1s {\footnotesize NetVLAD} matching + 1s VGGT inference + $\sim$3 s refinement = $\sim$5 s total per submap pair.

While submap fusion is performed once exploration ends, collaboration itself is online: keyframe encodings are broadcast continuously, and each agent uses them to detect overlap with other trajectories and to locally increase its keyframe density in those regions. We deliberately avoid tightly-coupled multi-agent bundle adjustment: its bandwidth requirements are incompatible with the degraded-network regime we target. \cref{tab:dyn_kf} isolates this contribution. On Sq1\&FastStr the gain is limited, as every FastStr keyframe is already covisible and aligned; on the two other configurations, disabling dynamic keyframing degrades global ATE by 43\% and 72\%.
\begin{table}[h]
  \centering
  \resizebox{0.8\columnwidth}{!}{%
  \setlength{\tabcolsep}{3pt}
  \begin{tabular}{lcc}
\hline
Scenes & with Dynamic KF & without Dynamic KF \\
    \midrule
    Sq1\&EgoCtr & 37.33 & 53.40 \\
    Sq1\&FastStr & 18.12 &  19.08\\
    FastStr\&Sq1\&EC & 39.63 &  68.13 \\ \hline
  \end{tabular}
  }
  \caption{Dynamic KF impact on ATE (cm) on global map. Enabling it brings improved alignment on all scenes. It has limited contribution on Sq1\&FastStr as the second trajectory is rectilinear and forward, thus all the keyframes carry the same information}
  \label{tab:dyn_kf}
\end{table}
\section{Conclusion}
We presented a novel Gaussian Splatting-based SLAM system that leverages a metric monocular depth estimation model (Depth Pro) to construct scale-consistent 3D models without requiring dedicated depth sensors, such as LiDAR, which are typically unavailable on consumer-grade hardware. To further enhance tracking robustness and accelerate convergence, we integrate IMU data and show its contribution. Additionally, we introduce an approach for rapid submap alignment using VGGT, enabling collaborative mapping. Our method demonstrates improvements over state-of-the-art approaches, achieving both superior reconstruction quality and reduced tracking error compared to other monocular 3DGS methods, as well as comparable quality to RGB-D systems while relying only on RGB and inertial data. In future work, we aim to employ a single model for both depth estimation and camera pose estimation and reduce its computational requirements to enable fully onboard execution on consumer smartphones, paving the way for a completely integrated SLAM solution.
\newpage


%
%

\bibliographystyle{splncs04}
\bibliography{main,references/references,references/collaborative_slam,references/deep_slam,references/export_review,references/gs_slam,references/vi_slam,references/others,references/datasets,references/models}

\clearpage
\appendix
\setcounter{section}{0}
\renewcommand{\thesection}{\Alph{section}} 
\renewcommand{\thesubsection}{\Alph{section}.\arabic{subsection}}

\setcounter{equation}{0}
\setcounter{figure}{0}
\setcounter{table}{0}
\setcounter{page}{1}

\makeatletter
\renewcommand{\theequation}{S\arabic{equation}}
\renewcommand{\thefigure}{S\arabic{figure}}
\renewcommand{\thetable}{S\arabic{table}}
\makeatother

\begin{center}
\textbf{\large Supplementary Material: CGS-SLAM — Collaborative Gaussian Splatting based SLAM for Multi-Agent Reconstruction}
\end{center}

\section{Representation: 3DGS}\label{app:3dgs}

This section details the Gaussian splatting technique used in our pipeline, it follows the original implementation by Kerbl \etal\cite{kerbl3Dgaussians}.
The underlying 3D scene is modeled using a collection of parametric Gaussian distributions. Each Gaussian model is denoted $ G = \{G_i\}_{i=1}^N $ and is characterized by the following parameters: a 3D position  $ \mu_i \in \mathbb{R}^3 $, a rotation and scaling matrix $R$ and $S$, representing the covariance matrix $ \Sigma_i \in \mathbb{R}^{3 \times 3} $ through the eigenvalue decomposition, an opacity value $ \alpha_i \in [0, 1] $ and a color vector $ c_i \in \mathbb{R}^3 $.

The probability density function of a Gaussian centered at $ \mu $ with covariance $ \Sigma $ is given by:
\begin{equation}
    G(x) = e ^{ -\frac{1}{2} (x - \mu)^T \Sigma^{-1} (x - \mu)},
\end{equation}
where $ x \in \mathbb{R}^3 $ represents a spatial point.

To maintain the symmetry and positive semi-definiteness of $ \Sigma $ during optimization, an eigenvalue decomposition is employed. This allows to express $ \Sigma $ as:
\begin{equation}
    \Sigma = R S S^T R^T,
\end{equation}
where $ R $ is an orthonormal rotation matrix and $ S $ is a diagonal matrix of scaling factors. This decomposition ensures numerical stability and physical plausibility of the Gaussian shapes.

To render the scene, the 3D Gaussians are projected onto the 2D image plane using a process known as \emph{splatting}. The resulting 2D covariance matrix $ \Sigma' $ is computed as:
\begin{equation}
    \Sigma' = J W \Sigma W^T J^T,
\end{equation}
where $ W $ is the world-to-view transformation matrix, and $ J $ is the Jacobian of the linearized projection mapping.

Given a set of Gaussians $ G $ and a camera pose $ T_c $, the color $ C $ of a pixel is computed by blending the contributions of all Gaussians that project onto the pixel, ordered by increasing depth. The blending formula is:
\begin{equation}
    C(G, T_c) = \sum_{i=1}^{N} c_i \alpha_i \prod_{j=1}^{i-1} (1 - \alpha_j),
\end{equation}
where $ c_i $ is the color of the $ i $-th Gaussian, and $ \alpha_i $ is the local opacity sampled from the 2D projection of that Gaussian at the pixel location.

The final pixel opacity $ O $ is defined analogously:
\begin{equation}
    O(G, T_c) = \sum_{i=1}^{N} \alpha_i \prod_{j=1}^{i-1} (1 - \alpha_j).
\end{equation}

Because the entire rendering process is differentiable with respect to the Gaussian parameters, we can optimize $ G $ using gradient-based methods. Specifically, we minimize the $ L_1 $ reconstruction loss between the rendered image $ C(G, T_c) $ and a reference ground-truth image $ I $:
\begin{equation}
    \mathcal{L}_{\text{photo}} = \| I - C(G, T_c) \|_1.
\end{equation}

\section{Message passing between clients and server}\label{app:message_passing}

\begin{figure}[htbp]
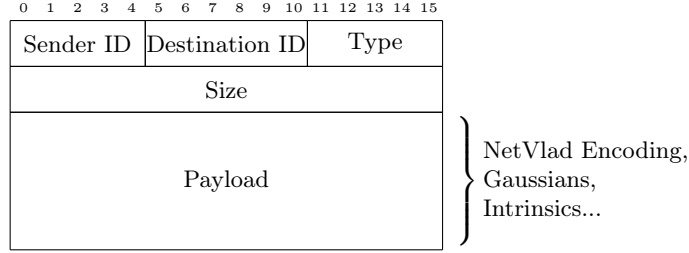

\centering
\begin{bytefield}[bitwidth=1.1em]{16}
\bitheader{0-15} \\
\bitbox{5}{Sender ID} & \bitbox{6}{Destination ID} & \bitbox{5}{Type}\\
\wordbox{1}{Size} \\
\begin{rightwordgroup}{NetVlad Encoding, \\Gaussians, \\Intrinsics...}
\wordbox{3}{Payload}
\end{rightwordgroup} \\
\end{bytefield}
\caption{Our communication protocol uses a custom message architecture enabling efficient, low-bandwidth communication of keyframe encodings.}
\label{fig:packet_structure}
\end{figure}

To enhance the selection of keyframe candidates for the alignment process described in \cref{align}, we introduce a lightweight, scalable message-passing framework that enables inter-agent coordination without requiring centralized mapping or full trajectory sharing. The server operates in an idle state until it receives initialization messages from individual agents, each of which begins by transmitting its intrinsic camera parameters for potential heterogeneity in sensor models across clients. Once initialized, the server assumes a broadcast role: it collects and caches compact keyframe encodings from all active agents, then disseminates them in real time to all connected clients. This enables each agent to continuously monitor for spatial overlaps with other agents’ trajectories by comparing incoming encodings against its own local keyframe database. 
As the keyframe encoder, we adopt NetVLAD~\cite{netvlad}, a lightweight, descriptor-based architecture widely validated for place recognition and visual localization tasks. No incorrect submap association occurred on the datasets evaluated here, but robustness in visually repetitive environments remains to be quantified. Future work will consider stronger encoders such as DINOv3~\cite{dinov3}, following \cite{MAGIC-SLAM} and its use of DINOv2 encodings~\cite{dinov2}.
To ensure low-latency communication and minimal bandwidth overhead, we have designed a compact, extensible message protocol. The architecture supports multiple message types, including initialization, keyframe encoding, pose updates, and alignment triggers while maintaining a minimal fixed-size header and variable-length payload. The message layout is shown in \cref{fig:packet_structure}.

\section{IMU pre-integration}\label{app:imu}
Most consumer-grade devices embed 6-degree-of-freedom (6-DOF) IMUs which include accelerometers and gyroscopes that measure linear acceleration, $ \mathbf{a} = [\ddot{x}, \ddot{y}, \ddot{z}] $, and angular velocity, $ \dot{\Theta} = [\dot{\alpha}, \dot{\beta}, \dot{\gamma}] $, in 3D space.
The displacement between two consecutive time steps, $ \Delta \mathbf{p}_{t}^{t-1} $, expressed in the coordinate frame of the previous time step, can be computed as:
\begin{equation}
    \Delta \mathbf{p}_{t}^{t-1} = \mathbf{v}_{t-1} \cdot \Delta t + \frac{1}{2} \mathbf{a}_t \cdot (\Delta t)^2,
\end{equation}
where the velocity $ \mathbf{v}_{t-1} $ is obtained by numerically integrating the previous acceleration over time:
\begin{equation}
    \mathbf{v}_{t-1} = \mathbf{v}_{t-2} + \mathbf{a}_{t-1} \cdot \Delta t.
\end{equation}

Similarly, the angular displacement $ \Delta \Theta_t^{t-1} $, again expressed in the previous frame, is given by:
\begin{equation}
    \Delta \Theta_t^{t-1} = \dot{\Theta}_{t-1} \cdot \Delta t.
\end{equation}

Using these estimates, the relative transformation $ {}^{t-1}_t \mathbf{T}_I $ between the two IMU frames can be constructed as:
\begin{equation}\label{fusion_equation}
    {}^{t-1}_t \mathbf{T}_I = \left[ {}^{t-1}_t \mathbf{R} \mid \Delta \mathbf{p}_t^{t-1} \right],
\end{equation}
where $ {}^{t-1}_t \mathbf{R} $ is a rotation matrix derived from the angular displacement $ \Delta \Theta_t^{t-1} $.

To express this transformation in the camera coordinate frame, we apply the static transformation $ {}^C_I \mathbf{T} $ between the IMU and the camera, yielding the relative camera transformation:
\begin{equation}
    {}^{t-1}_t \mathbf{T}_C = {}^C_I \mathbf{T} \cdot {}^{t-1}_t \mathbf{T}_I.
\end{equation}

\section{Implementation Details}

\Cref{tab:parameters} lists the specific hyperparameters and parameters used throughout our method for training, tracking, and global mapping. The mapping process uses a set of balanced loss weights ($\lambda_D, \lambda_C, \lambda_S$) and densification intervals to maintain a high-fidelity representation of the scene while preventing Gaussian explosion. For tracking, we enforce a strict pixel opacity threshold and covisibility constraints to ensure frame-to-frame pose accuracy. Finally, the global consistency parameters, including dedicated learning rates for global position and rotation, are used during the optimization phase of the submap alignment process.
\begin{table}[ht]
    \centering

    \begin{minipage}{0.48\textwidth}
        \centering
        \begin{tabular}{lr}
            \toprule
            
            \textbf{Mapping} & \textbf{Value} \\
            \midrule
            \textit{Learning Rates (LR)} & \\
            \(\lambda_D \) & 0.05 \\
            \(\lambda_C \) & 0.75 \\
            \(\lambda_S \) & 0.2 \\
            Camera $t$ / $q$ LR & $0.001$ / $0.003$ \\
            Feature / Opacity LR & $0.0025$ / $0.05$ \\
            Scaling / Rotation LR & $0.001$ / $0.001$ \\
            \addlinespace
            \textit{Densification} & \\
            Densification interval & $40$ \\
            Pruning interval & $50$ \\
            Max steps & $30,000$ \\
            \bottomrule
        \end{tabular}
    \end{minipage}
    \hfill
    \begin{minipage}{0.48\textwidth}
        \centering
        \begin{tabular}{lr}
            \toprule
            \textbf{Keyframing} & \textbf{Value} \\
            \midrule
            Covisibility threshold & $0.9$ \\
            Min KF \(s\) & $2$ \\
            Max KF \(k\) & $5$ \\
            \addlinespace
            \midrule
            \textbf{Tracking} & \textbf{Value} \\
            \midrule
            \(\lambda_D \) & $0.001$ \\
            Pixel opacity threshold & $0.99$ \\
            \addlinespace
            \toprule
            \textbf{Global Consistency} & \textbf{Value} \\
            \midrule
            Translation LR (global) & $0.01$ \\
            Rotation LR (global) & $0.01$ \\
            Neighbor frames & $3$ \\
            Iterations & $300$ \\
            \addlinespace
            \bottomrule
        \end{tabular}
    \end{minipage}
    \small 
    \caption{System parameters used in our experimental setup.}
    \label{tab:parameters}
\end{table}
\section{Two-Stage Procrustes Alignment}\label{app:alignement}
Given a pair of matching keyframes $(f_1, f_2)$ from clients 1 and 2 respectively, we collect their temporal neighbors $\mathcal{N}_1, \mathcal{N}_2$ to form two point sets. For robustness to rotation ambiguity, we augment each point set with a virtual point offset along the camera's up-axis:
\begin{equation}
\mathbf{p}^+_i = \mathbf{p}_i + \lambda \cdot \mathbf{u}_i,
\end{equation}
where $\mathbf{p}_i \in \mathbb{R}^3$ is the camera position, $\mathbf{u}_i$ is the local $y$-axis (up-vector) of the camera frame, and $\lambda$ is set to the mean inter-camera distance within the point set. Let $\mathcal{P}_k = \{\mathbf{p}_i\}_{i \in \{f_k\} \cup \mathcal{N}_k} \cup \{\mathbf{p}^+_{f_k}\}$ denote the augmented point set for client $k$.

\textbf{Stage 1: VGGT $\to$ Sub-map 1.} We estimate a similarity transform $\mathbf{T}_1 \in Sim(3)$ that aligns the VGGT camera positions of client 1's keyframes to their corresponding sub-map poses. Denoting by $\mathbf{t}(\mathbf{M})$ the translation component of a $4\times 4$ matrix $\mathbf{M}$, we solve:
\begin{equation}
\mathbf{T}_1 = \underset{\mathbf{T} \in Sim(3)}{\arg\min} \sum_{\mathbf{p} \in \mathcal{P}_1^{\text{vggt}}} \left\| \mathbf{T} \cdot \mathbf{p} - \tilde{\mathbf{p}} \right\|^2,
\end{equation}
where $\mathcal{P}_1^{\text{vggt}}$ contains positions extracted from inverted VGGT extrinsics $\{\mathbf{t}(\mathbf{E}_i^{-1})\}$ and $\tilde{\mathbf{p}}$ denotes the corresponding sub-map position $\{\mathbf{t}((\mathbf{T}^{S_1}_{C_i})^{-1})\}$.

\textbf{Stage 2: Transformed VGGT $\to$ Sub-map 2.} We apply $\mathbf{T}_1$ to client 2's VGGT poses, bringing them into the coordinate frame of sub-map 1. We then estimate $\mathbf{T}_2$ aligning these transformed poses to sub-map 2:
\begin{equation}
\mathbf{T}_2 = \underset{\mathbf{T} \in Sim(3)}{\arg\min} \sum_{\mathbf{p} \in \mathcal{P}_2^{\text{vggt}}} \left\| \mathbf{T} \cdot (\mathbf{T}_1 \cdot \mathbf{p}) - \tilde{\mathbf{p}} \right\|^2,
\end{equation}
where $\tilde{\mathbf{p}}$ now refers to sub-map 2 positions $\{\mathbf{t}((\mathbf{T}^{S_2}_{C_j})^{-1})\}$.

Both stages are solved via Procrustes analysis and subsequently refined by gradient descent on the full pose error, including orientation. Since $\mathbf{T}_1$ maps VGGT coordinates to sub-map 1 and $\mathbf{T}_2$ maps sub-map 1 coordinates (via the transformed VGGT poses) to sub-map 2, the composition $\mathbf{T}_2 \circ \mathbf{T}_1$ effectively transforms from VGGT to sub-map 2. The alignment from sub-map 2 to sub-map 1 is therefore:
\begin{equation}
\boxed{\mathbf{T}^{S_1}_{S_2} = \mathbf{T}_2^{-1}}
\end{equation}

\section{Experiments with Depth Anything V3}\label{app:da3}
\begin{figure}
    \centering
    \includegraphics[width=1\linewidth]{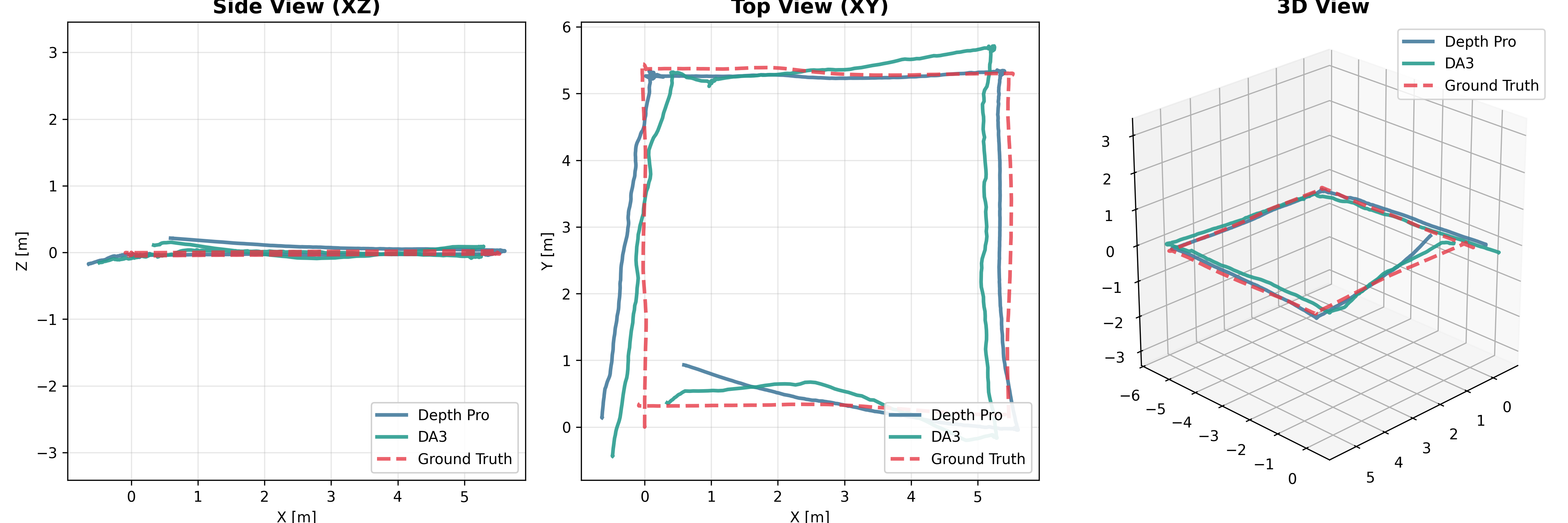}
    \caption{Trajectory comparison between Depth Pro and Depth Anything configuration. We can see that the Depth Pro configuration achieves better tracking, with better handling of turns. }
    \label{fig:trajda3}
\end{figure}

In the aim of reducing the dependency on two deep neural network models, we tested our method on Depth Anything V3\cite{depthanything3} (DA3) which produces both metric depth maps and camera pose estimation. It could then replace Depth Pro for the MMDE part of the local mapping and VGGT for submap alignment in the global consistency. \cref{tab:da3vsdp} reports the comparison, and \cref{fig:trajda3} a corresponding trajectory.
However, our results showed lower quality for the local mapping part, and thus contamination of the submap alignment process, which made us choose to keep the dual model approach.
\begin{table}[h]
  \centering

  \setlength{\tabcolsep}{1.5pt}
  \begin{tabular}{lcccccccc}
   & \multicolumn{4}{c}{Depth Pro} & \multicolumn{4}{c}{Depth Anything V3} \\
  \cmidrule(lr){2-5} \cmidrule(lr){6-9} 
 Scene&ATE-RMSE&PSNR&SSIM&LPIPS&ATE-RMSE&PSNR&SSIM&LPIPS\\
    \midrule
        Square-1 & 42.33&18.13&0.64&0.43 & 61.42& 16.42& 0.53&0.45 \\        
        Square-2 & 55.24&18.08&0.62 & 0.39& X&X&X&X\\
        Ego-centric & 4.26&23.34&0.86&0.20&26.55 & 18.62& 0.69& 0.35 \\
        Fast-Straight & 6.07 &24.51&0.82&0.26 &12.63& 20.54&0.68&0.48\\
        Slow-Straight-2 & 5.01 &25.85&0.84&0.24 &6.18& 25.66&0.84&0.24\\
            \midrule

        fr1/desk & 4.2 & 19.44 & 0.70& 0.38& 3.26&21.83&0.77&0.29\\
        fr1/desk2 & 5.12 & 18.53 & 0.62&  0.43 &7.83&20.18&0.72&0.36 \\
        fr1/room & 5.3 &15.41& 0.52&  0.48 &10.36& 14.84&0.48&0.49\\
        fr2/xyz & 1.14 & 23.24 & 0.84& 0.20 & 1.32 & 25.40&0.84&0.19 \\
        fr3/office & 20.42 & 18.39 &0.67& 0.42 & 33.82 & 18.32&0.63&0.46\\
        
    \end{tabular}
    
   \caption{Tracking (ATE-RMSE $\downarrow$) and Rendering (PSNR \(\uparrow\), SSIM\(\uparrow\) and LPIPS\(\downarrow\)) comparison between Depth Pro (DP) and Depth Anything V3 (DA3) setup on TUM and UT-MM datasets. DA3 is competitive and occasionally better on the small TUM sequences, but degrades sharply on the large UT-MM scenes and fails on Square-2, which is why we retain the dual-model setup}
   \label{tab:da3vsdp}
\end{table}
\section{Tests on Replica Multiagent Dataset with the constant velocity hypothesis}

Replica Multiagent~\cite{replica19arxiv,cp-slam} provides no inertial measurements, so running CGS-SLAM on it requires replacing the IMU prior with a constant-velocity assumption.
\Cref{tab:imu} quantifies the cost of this substitution on UT-MM (7.05\,cm of average ATE). The results below are reported for completeness and are not directly comparable to the main evaluation.

\begin{figure}
    \centering
    \includegraphics[width=0.8\linewidth]{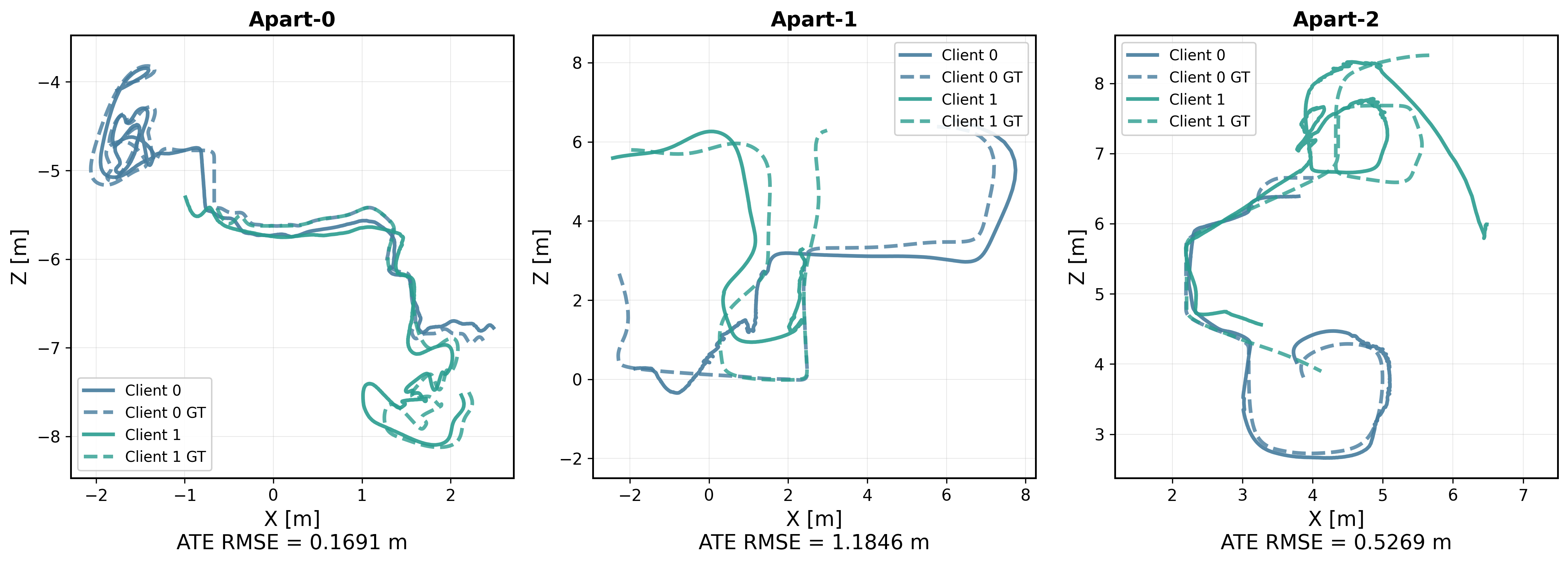}
    \caption{Top view of trajectories from global alignment of submaps from ReplicaMultiagent. The tracking error from the lack of IMU contaminates the submap alignments, resulting in poor ATE-RMSE.}
    \label{fig:replica_trio_traj}
\end{figure}

\begin{figure}[h]
    \centering
    \includegraphics[width=\linewidth]{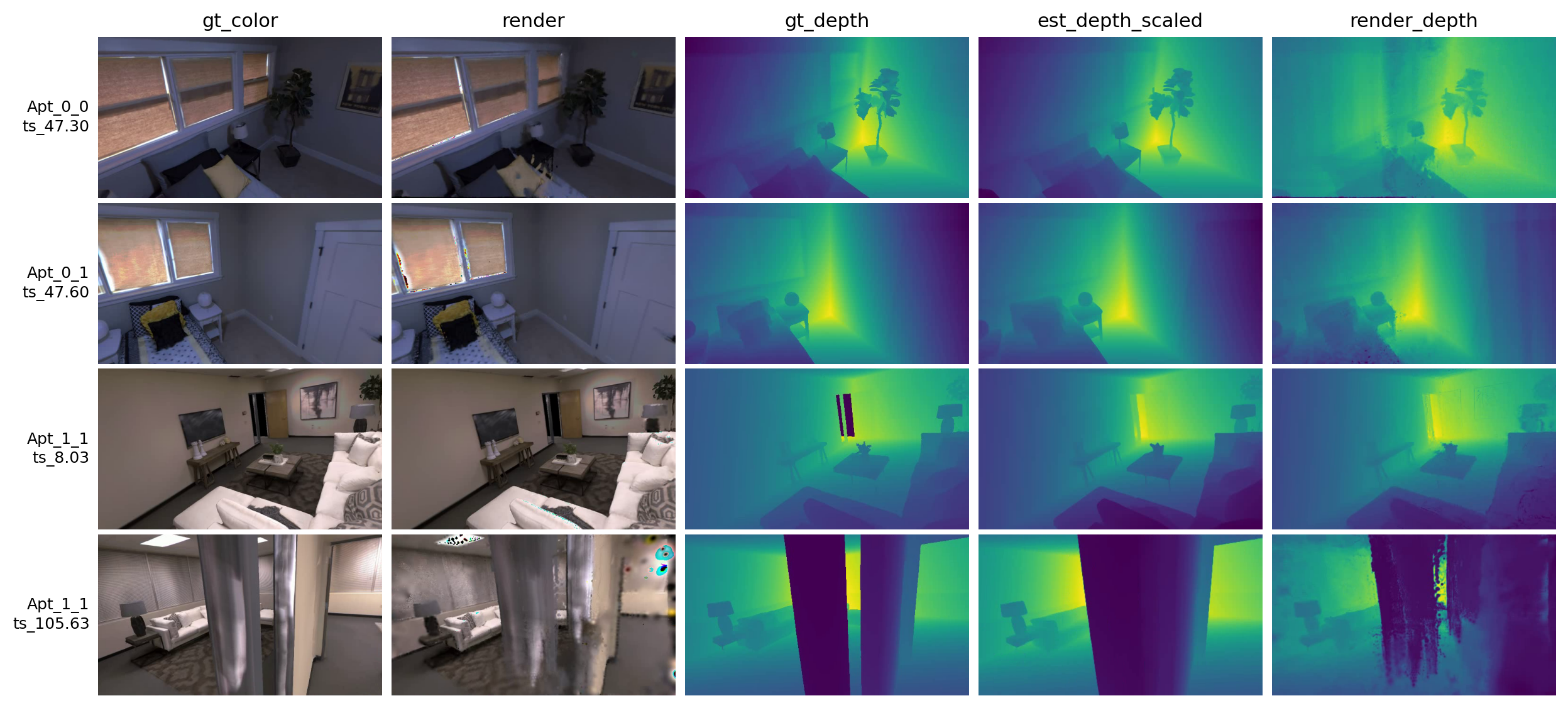}
    \caption{Rendering examples of the scenes from Apt-0 and Apt-1 of the Replica dataset.
}
    \label{fig:replica_render}
\end{figure}

As reported in \cref{tab:replica}, results are bimodal rather than uniformly degraded. Apt-0 and Apt-2/agent-0 stay within the range of our IMU-based results, while Apt-1 and Apt-2/agent-1 diverge above one meter (\cref{tab:replica,fig:replica_trio_traj}). Since both agents of Apt-2 observe the same scene, the separation follows the trajectories rather than the environment. The diverging runs are those crossing between rooms, where a doorframe passes within a few centimeters of the camera. Our depth estimation is unreliable at such close range on synthetic imagery, and the resulting keyframe injects grossly mis-scaled geometry into the Gaussian map; without an inertial prior to constrain the following poses, tracking has no way to recover. An example of this misprediction is visible on the last line of \cref{fig:replica_render}. The failure is therefore driven by near-field depth estimation rather than by the collaborative back-end, and it is confined to a component our pipeline treats as interchangeable (\Cref{app:da3}).

\begin{table}[h]
  \centering

  \setlength{\tabcolsep}{1.5pt}
  \begin{tabular}{lccccc}

 Scene&Agent&ATE-RMSE&PSNR&SSIM&LPIPS\\
    \midrule
        \multirow{2}{*}{Apt-0} & 0 & 12.33&26.96&0.79 &  0.29 \\
                             & 1 & 15.06&25.20&0.81 &  0.24  \\
                                         \midrule
        \multirow{2}{*}{Apt-1} & 0 & 199.39&12.92&0.55&0.49\\
                             & 1 &122.20 &17.65&0.62&0.43\\
                                         \midrule
        \multirow{2}{*}{Apt-2} & 0 & 19.54&25.20&0.77 & 0.28 \\
                             & 1 & 150.09&18.83&0.72 & 0.34 \\
                                         \bottomrule
        
    \end{tabular}
    
   \caption{Tracking (ATE-RMSE $\downarrow$cm) and Rendering (PSNR \(\uparrow\)dB, SSIM\(\uparrow\) and LPIPS \(\downarrow\)) on Replica Dataset under the constant velocity hypothesis (no IMU).}
   \label{tab:replica}
\end{table}
\enlargethispage{2\baselineskip}

\section{Supplementary Results}
\begin{figure}[h]
    \centering
    \includegraphics[width=\linewidth]{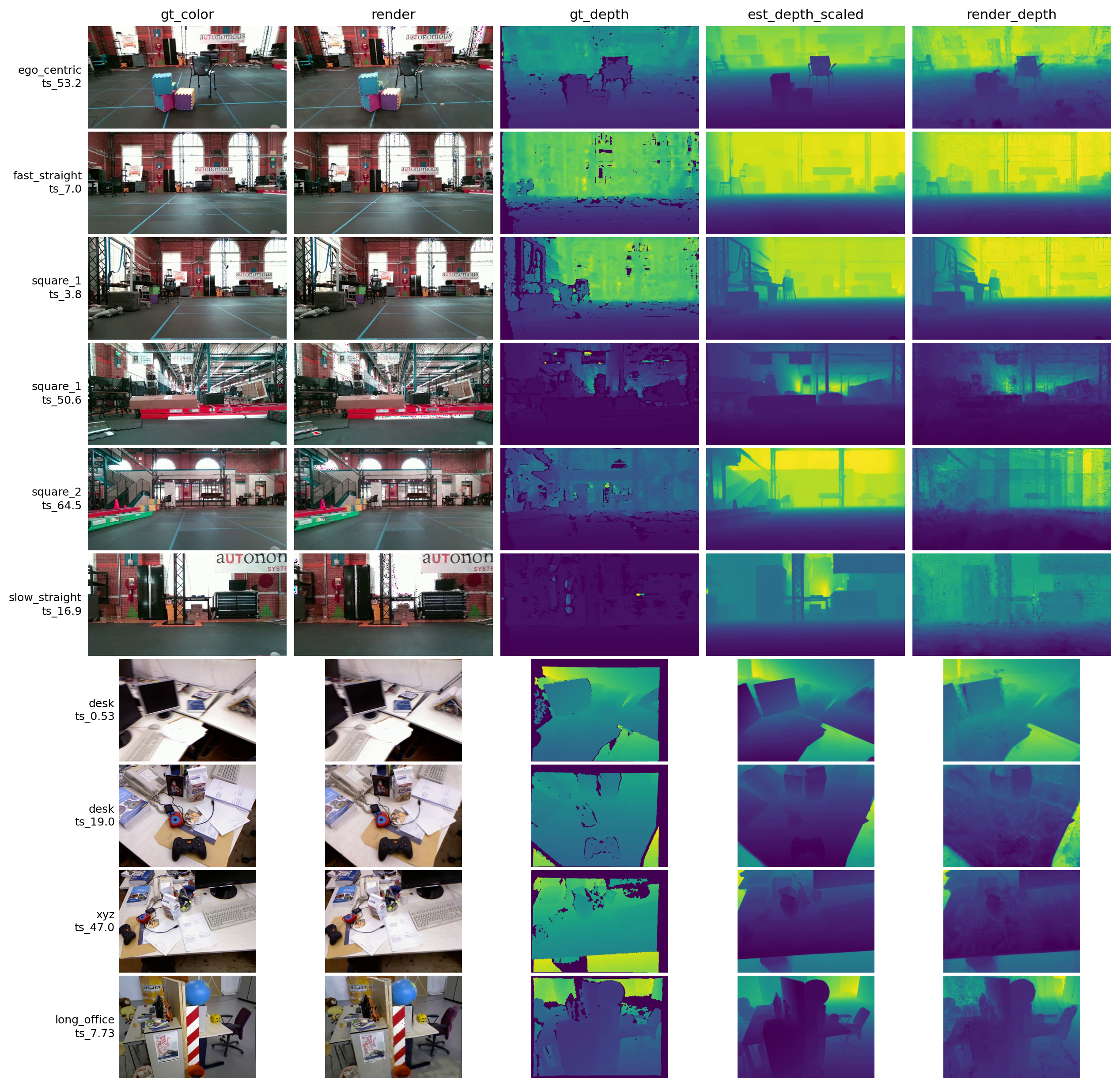}
    \caption{Rendering results of monocular single agent exploration on UT-MM and TUM dataset with Depth Pro as MMDE. We can see that the rendering provides good results, but also some small artifacts. Leveraging a MMDE allows to bypass some of the poorly detailed area of the depth ground truth from UT-MM.}
    \label{fig:renderingmore}
\end{figure}

For completeness, tracking results of Magic-SLAM and MAC-Ego3D on the full UT-MM dataset are reported in \cref{app:utmmate}. We also provide supplementary rendering results in \cref{fig:renderingmore}

\begin{table}
\centering
\resizebox{\columnwidth}{!}{%
\begin{tabular}{lccccc}
  Method & Config & Square-1 & Square-2 & Ego-Centric-1 & Fast-Straight\\
  \midrule
  \footnotesize{Magic-SLAM}&RGBD & \best{6.24} & X & 4.78 & \best{5.72} \\
  \footnotesize{MAC-Ego3D}&RGBD & 40.07 & 77.24 & 39.08 & 19.44 \\
  \footnotesize{Ours} &RGB+IMU &42.33 & 55.24 & \best{4.26} & 6.07\\
  \midrule
   &    & Slow-Straight-1 & Slow-Straight-2 & Ego-Drive & Ego-Centric-2\\
  \midrule
  \footnotesize{Magic-SLAM}&RGBD & 8.49 & 5.94 & 8.01 & 7.16 \\
  \footnotesize{MAC-Ego3D}&RGBD & 13.82 & 16.02 & X & 19.31 \\
  \footnotesize{Ours} &RGB+IMU &\best{1.43} & \best{5.01} & 15.86 & \best{5.50}\\
  \end{tabular}

   } 
   \caption{\footnotesize{Single agent tracking evaluation ATE (cm). Ours (RGB) against Magic-SLAM (RGB-D) and MAC-Ego3D (RGB-D) on UT-MM scenes.} }
   \label{app:utmmate}
   
   \end{table}

\cref{app:utmmate} extends the single-agent evaluation to the full UT-MM dataset. Ours is the only system to complete all eight sequences: Magic-SLAM diverges on Square-2 and MAC-Ego3D on Ego-Drive. Performance is governed by the motion profile of each sequence. On the straight and localized sequences we stay in the range of the RGB-D baselines without using a depth sensor. The Square sequences, which chain four sharp turns, are the exception, and they are difficult for every evaluated system (MAC-Ego3D: 40.07/77.24 cm; MM3DGS: 47.05/68.50 cm, \cref{tab:imu}). Sharp turns provide little translational parallax, so heading error accumulates; Magic-SLAM recovers on Square-1 through loop closure and global bundle adjustment, which we omit for bandwidth reasons. Robustness to rotation-dominated motion is the main avenue for improvement.

\end{document}